\documentclass[11pt,a4paper]{article}
\usepackage[hyperref]{acl2020} 
\usepackage{times}
\usepackage{latexsym}

\usepackage{microtype}

\aclfinalcopy 
\def\aclpaperid{2728} 

\usepackage{xspace,mfirstuc,tabulary}
\usepackage{breakurl}

\author{
 Pratik Jawanpuria, \ Mayank Meghwanshi, \ Bamdev Mishra\\
 Microsoft, India\\
 \texttt{\{pratik.jawanpuria,mamegh,bamdevm\}@microsoft.com}\\
}

\date{}

\usepackage{enumitem}
\setlist{nolistsep}
\usepackage{nicefrac}

\usepackage{booktabs}
\usepackage{mathtools}
\usepackage{color,multirow,wrapfig}
\usepackage[normalem]{ulem}

\usepackage{graphicx} 
\usepackage{epsfig} 
\usepackage{subfigure} 
\usepackage{caption}

\usepackage{algorithm}
\usepackage{algorithmic}

\usepackage{amssymb}
\usepackage{amsmath}
\usepackage{amsthm}
\usepackage{mathtools}

\newcommand{\thead}[1]{\multicolumn{1}{c}{\textbf{#1}}}
\newcommand{\lhead}[1]{\multicolumn{1}{l}{\textbf{#1}}} 

\newcommand{\algname}{\mbox{MBA}}

\newcommand{\mat}[1]{{\bf #1}}
\def\R{\mathbb{R}}
\newcommand{\norm}[1]{\left\|#1\right\|}

\def\bA{\mathbf A}

\def\bC{\mathbf C}
\def\bD{\mathbf D}
\def\bI{{\mathbf I}}

\def\bU{{\mathbf U}}
\def\bV{{\mathbf V}}
\def\bW{{\mathbf W}}
\def\bX{{\mathbf X}}
\def\bY{{\mathbf Y}}
\def\bZ{{\mathbf Z}}

\def\DS{{\mathbb{DS}}}

\def\minop{\mathop{\rm min}\limits}

\title{Geometry-aware Domain Adaptation for\\ Unsupervised Alignment of Word Embeddings}
\begin{document} 
\maketitle

\begin{abstract} 
We propose a novel manifold based geometric approach for learning unsupervised alignment of word embeddings between the source and the target languages. Our approach formulates the alignment learning problem as a domain adaptation problem over the manifold of doubly stochastic matrices. 
This viewpoint arises from the aim to align the second order information of the two language spaces. 
The rich geometry of the doubly stochastic manifold allows to employ efficient Riemannian conjugate gradient algorithm for the proposed formulation.
Empirically, the proposed approach outperforms state-of-the-art optimal transport based approach on the bilingual lexicon induction task across several language pairs. The performance improvement is more significant for distant language pairs.
\end{abstract}

\section{Introduction}
Learning bilingual word embeddings is an important problem in natural language processing  \citep{mikolov13a,faruqui14a,artetxe16a,conneau18a}, with usage in cross-lingual information retrieval \citep{vulic15a}, text classification \citep{wan11a,klementiev12a}, machine translation \citep{artetxe18c}  etc. 
Given a source-target language pair, the aim is to represent the words in both languages in a common embedding space. This is usually achieved by learning a linear function that maps word embeddings of one language to the embedding space of the other language \citep{mikolov13a}. 

Several works have focused on learning such bilingual mapping in supervised setting, using a bilingual dictionary during the training phase  \citep{artetxe18a,joulin18b,mjaw19a}. Recently, unsupervised bilingual word embeddings have also been explored  \citep{zhang17a,zhang17b,conneau18a,artetxe18b,hoshen18,grave18,alvares18a,zhou19a,mjaw20a}. 



Learning unsupervised cross-lingual mapping may be viewed as an instance of the more general unsupervised domain adaptation problem \citep{ben07a,gopalan11a,sun16a,mahadevan18a}. The latter fundamentally aims at aligning the input feature (embeddings) distributions of the source and target domains  (languages). 
In this paper, we take this point of view and learn cross-lingual word alignment by finding alignment between the second order statistics of the source and the target language embedding space. 

We formulate a novel optimization problem on the set of doubly stochastic matrices. The objective function consists of matching covariances of words from source to target languages in a least-squares sense. For optimization, we exploit the fact that the set of doubly stochastic matrices has rich geometry and forms a Riemannian manifold \citep{douik19a}. The Riemannian optimization framework \citep{absil08a,edelman98a,smith94a} allows to propose a computationally efficient conjugate gradient algorithm \citep{douik19a}. Experiments show the efficacy of the proposed approach on the bilingual lexicon induction benchmark, especially on the  language pairs involving distant languages. 


\section{Motivation and Related Work} \label{sec:related_work}
We introduce the bilingual word alignment setup followed by a discussion on domain adaptation approaches.

\noindent\textbf{Bilingual alignment.} Let $\bX\in\R^{n\times d}$ and $\bZ\in\R^{n\times d}$ be $d$-dimensional word embeddings of $n$ words of the source and the target languages, respectively. The aim is to learn a linear operator $\bW: \R^d\rightarrow\R^d$ that best approximates source embeddings in the target  language space. 

In the supervised setup, a list of source words and their translations in the target language is provided. This is represented by an \emph{alignment} matrix $\bY$ of size $n\times n$, where $\bY_{ij}=1$ if $j$-th word in the target language is a translation of the $i$-th word in the source language and $\bY_{ij}=0$ otherwise. 
A standard way to learn orthogonal $\bW$ is by solving the \textit{orthogonal Procrustes} problem \citep{artetxe16a,smith17a}, i.e., 
\begin{equation} \label{eq:procrustes}
\begin{array}{lll}
\minop_{\bW \in\R^{d\times d}} & \norm{\bX \bW - \bY \bZ}_{\rm Fro}^2\\
{\rm subject\ to} & \bW^\top \bW = \bI,
\end{array}
\end{equation}
where $\norm{\cdot}_{\rm Fro}$ is the Frobenius norm and $\bI$ is the identity matrix. Problem (\ref{eq:procrustes}) has the closed-form solution $\bW^\star = \bU\bV^\top$, where $\bU$ and $\bV$ are the respective left and right orthogonal factors of the singular value decomposition of $\bX^\top \bY \bZ$ \citep{schonemann1966a}. 

In the unsupervised setting, $\bY$ is additionally unknown apart from $\bW$. Most unsupervised works \citep{zhang17b,artetxe18b,grave18,conneau18a} tackle this challenge by learning $\bY$ and $\bW$ jointly. However, their performance rely on finding a good initialization candidate for the alignment matrix $\bY$ \cite{zhang17b,grave18,alaux19a,mjaw20a}. 



Performing optimization over the set of binary matrices, $\bY\in\{0,1\}^{n\times n}$, to learn the bilingual alignment matrix is computationally hard. Hence, some    works \citep{zhang17b,xu18a} view the source and the target word embedding spaces as two distributions and learn $\bY$ as the transformation that makes the two distributions close. 
This viewpoint is based on the theory of \emph{optimal transport} \citep{villani09,peyre19a}. $\bY$ is, thus, modeled as a \emph{doubly stochastic} matrix: the entries in $\bY \in [0,1]$ and each row/column sums to $1$. 
Permutation matrices are extreme points in the space of doubly stochastic matrices.

\citet{alvares18a} propose learning the doubly stochastic $\bY$ as a transport map between the \emph{metric spaces} of the words in the source and the target languages. 
They optimize the {Gromov-Wasserstein} (GW) distance, which measures how distances between pairs of words are mapped across languages.
For learning $\bY$, they propose to
\begin{equation}\label{eq:GW}
\begin{array}{lll}
\minop_{\bY \in \DS^n} & - {\rm Trace}(\bY^\top \bC_\bX \bY \bC_\bZ) ,
\end{array}
\end{equation}
where $\DS^n \coloneqq \{ \bY \in \R^{n\times n} : \bY \geq 0,  \bY^\top \mat{1} = \mat{1} \text{ and } \bY \mat{1} = \mat{1} \} $ is the set of $n\times n$ doubly stochastic matrices, $\bY \geq 0$ implies entry-wise non-negativity, $\mat{1}$ is a column vector of ones, and $\bC_\bX = \bX \bX^\top$ and $\bC_\bZ = \bZ \bZ^\top$ are $n\times n$ word covariance matrices of source and target languages, respectively. An iterative scheme is proposed for solving (\ref{eq:GW}), where each iteration involves solving an optimal transport problem with \emph{entropic regularization} \citep{peyre16a,peyre19a}. The optimal transport problem is solved with the popular \emph{Sinkhorn} algorithm \citep{cuturi13a}. 
It should be noted that the GW approach (\ref{eq:GW}) only learns $\bY$. The linear operator to map source language word embedding to the target language embedding space can then be learned by solving (\ref{eq:procrustes}). 



\noindent\textbf{Domain adaptation.} Domain adaption refers to transfer of information across domains and has been an independent research of interest in many fields including natural language processing \citep{daumeiii07a,borgwardt06a,adel17a,baktashmotlagh13a,fukumizu07a,wang15a,prettenhofer11a,wan11a,sun16a,mahadevan18a,ruder19a}.  

One modeling of interest is by \citet{sun16a}, who motivate a linear transformation on the features in source and target domains. In \citep{sun16a}, the linear map $\bA \in \R^{d\times d}$ is solved by 
\begin{equation}\label{eq:coral}
\begin{array}{ll}
\minop_{\bA \in \R^{d\times d}}	& \norm{\bA^\top \bD_\bX \bA - \bD_\bZ}_{\rm Fro}^2,
\end{array}
\end{equation}
where $\bD_1$ and $\bD_2$ are $d\times d$ are feature covariances of source and target domains (e.g., $\bD_\bX = \bX^\top \bX$ and $\bD_\bZ = \bZ^\top \bZ$), respectively. Interestingly, (\ref{eq:coral}) has a closed-form solution and shows good performance on standard benchmark domain adaptation tasks \citep{sun16a}.

\section{Domain Adaptation Based Cross-lingual Alignment}
The domain adaptation solution strategies of \citep{sun16a,mahadevan18a} can be motivated directly for the cross-lingual alignment problem by dealing with word covariances instead of feature covariances. However, the cross-lingual word alignment problem additionally has a bi-directional symmetry: if $\bY$ aligns $\bX$ to $\bZ$, then $\bY ^\top$ aligns $\bZ$ to $\bX$. 
We exploit this to propose a bi-directional domain adaptation scheme based on (\ref{eq:coral}). The key idea is to adapt the second order information of the source and the target languages into each other's domain. We formulate the above as follows: 
\begin{equation}\label{eq:proposed}
\begin{array}{lll}
\minop_{\bY \in \DS^{n}} &  \|  \bY^\top \bC_\bX \bY -  \bC_\bZ\|_{\rm Fro}^2 \\
& \quad + \ \ \|  \bY \bC_\bZ \bY^\top -  \bC_\bX\|_{\rm Fro}^2,
\end{array}
\end{equation}
%
The first term in the objective function $\|  \bY^\top \bC_\bX \bY -  \bC_\bZ\|_{\rm Fro}^2$ adapts the domain of $\bX$ (source) into $\bZ$ (target). 
Equivalently, minimizing only the first term in the objective function of (\ref{eq:proposed}) leads to row indices in $\bY^\top \bX$ aligning closely with the row indices of $\bZ$. 
Similarly, minimizing only the second term $\|  \bY \bC_\bZ \bY^\top -  \bC_\bX\|_{\rm Fro}^2$ adapts $\bZ$ (now treated as the source domain) into $\bX$ (now treated as the target domain), which means that the row indices $\bY\bZ$ and $\bX$ are closely aligned. Overall, minimizing both the terms of the objective function allows to learn the alignment matrix $\bY$ from $\bX$ to $\bZ$ and $\bY ^\top$ from $\bZ$ to $\bX$ simultaneously. Empirically, we observe that bi-directionality acts as a self regularization, leading to optimization stability and better generalization ability. 

The differences of the proposed formulation (\ref{eq:proposed}) with respect to the GW formulation (\ref{eq:GW}) are two fold. First, the formulation (\ref{eq:GW}) maximizes the inner product between $\bY^\top \bC_\bX \bY $ and $\bC_\bZ$. This inner product is sensitive to differences in the norms of $\bY^\top \bC_\bX \bY$ and $\bC_\bZ$. The proposed approach circumvents this issue since (\ref{eq:proposed})  explicitly penalizes entry-wise mismatch between $\bY^\top \bC_\bX \bY $ and $\bC_\bZ$. Second, the GW algorithm for (\ref{eq:GW}) is sensitive to choices of the entropic regularization parameter \citep{alvares18a,peyre19a}. In our case, no such regularization is required. 


Most recent works that solve optimal transport problem by optimizing over doubly stochastic matrices employ the Sinkhorn algorithm with entropic regularization \citep{cuturi13a,peyre16a,peyre19a}. 
In contrast, we exploit the Riemannian manifold structure of the set of doubly stochastic matrices ($\DS^n$) recently studied in \citep{douik19a}. $\DS^n$ is endowed with a smooth Fisher information metric (inner product) that makes the manifold smooth \citep{douik19a,sun15a,lebanon04a}. In differential geometric terms, $\DS^n$ has the structure of a Riemannian submanifold. This makes computation of optimization-related ingredients, e.g., gradient and Hessian of a function, projection operators, and retraction operator, straightforward. Leveraging the versatile Riemannian optimization framework \citep{absil08a,edelman98a,smith94a}, the constrained problem (\ref{eq:proposed}) is conceptually transformed to an \emph{unconstrained} problem over the nonlinear manifold. Consequently, most unconstrained optimization algorithms generalize well to manifolds. We solve (\ref{eq:proposed}) using the Riemannian conjugate gradient algorithm \citep{absil08a,douik19a}.

There exist several manifold optimization toolboxes such as Manopt \citep{boumal14a}, Pymanopt \citep{townsend16a}, Manopt.jl \citep{bergmann19a}, McTorch \citep{Meghwanshi_arXiv_2018} or ROPTLIB \citep{huang16a}, which have scalable off-the-shelf generic implementation of Riemannian algorithms. 
We use Manopt for our experiments, where we only need to provide the objective function (\ref{eq:proposed}) and its derivative with respect to $\bY$. The manifold optimization related ingredients are handled by Manopt internally. The computational cost per iteration of the algorithm is $O(n^2)$, which is similar to that of GW \citep{alvares18a}.

We term our algorithm as \textbf{M}anifold \textbf{B}ased \textbf{A}lignment (\algname) algorithm. Our code is available at \url{https://pratikjawanpuria.com/publications/}.

\begin{table*}[t]\centering
	{\small
		\centering
		\begin{tabular}{lrrrrrrrrrrrrr}
			\toprule
			\lhead{Method} & \thead{de-xx} & \thead{en-xx}  & \thead{es-xx} & \thead{fr-xx}  & \thead{it-xx}  & \thead{pt-xx} & \thead{xx-de} & \thead{xx-en}  & \thead{xx-es} & \thead{xx-fr}  & \thead{xx-it} & \thead{xx-pt} & \thead{avg.} \\
			\cmidrule(lr){1-1}
			\cmidrule(lr){2-7}
			\cmidrule(lr){8-13}
			\cmidrule(lr){14-14}
			GW            & 62.6 & 77.4 &\textbf{78.2} & \textbf{75.4} & \textbf{77.5}& 77.2 & 62.6 & 75.9 & \textbf{79.7} & \textbf{79.0} &\textbf{76.2} & 74.9 & 74.7 \\ 
			\textbf{\algname}    & \textbf{63.3} & \textbf{78.4}  & \textbf{78.2} & 75.3& 77.0& \textbf{77.5} & \textbf{63.1} & \textbf{77.3} & 79.4 & 78.7 & \textbf{76.2} &\textbf{75.0} & \textbf{75.0} \\
			\bottomrule
		\end{tabular}
	}
	\caption{P$@1$ for BLI on six European languages: English, German, Spanish, French, Italian, and Portuguese. Here `en-xx' refers to the average P$@1$ when English is the source language and others are target language. Similarly, `xx-en' implies English as the target language and others as source language. Thus, `avg.' shows P$@1$ averaged over all the thirty BLI results for each algorithm. The proposed algorithm {\algname} performs similar when the language pairs are closely related to each other.}\label{table:six-euro}
\end{table*}

\begin{table*}[h]\centering
	{\small
		\centering
		\begin{tabular}{lrrrrrrrrrrr}
			\toprule
			\lhead{Method} &  \thead{en-bg} & \thead{en-cs}  &  \thead{en-da} & \thead{en-el}& \thead{en-fi}&\thead{en-hu}&\thead{en-nl} & \thead{en-pl}  & \thead{en-ru} &     \\
			\cmidrule(lr){1-1}
			\cmidrule(lr){2-10}
				GW      &    22.8	 &  42.1	 &  54.4	 &  21.5 &  	37.7 &  	43.7 &  	72.9 &  	49.1 &  	36.1  &  	\\
			\textbf{\algname}  &  \textbf{38.1}  &  	\textbf{46.8}  &  	\textbf{56.1}  &  	\textbf{40.0} &  	\textbf{40.4}  &  	\textbf{46.1}  &  	\textbf{73.8}  &  	\textbf{50.4}  &  	\textbf{37.5}  &  	\\
			\midrule
			\lhead{Method} &   \thead{bg-en} & \thead{cs-en} & \thead{da-en}  & \thead{el-en} &  \thead{fi-en} &  \thead{hu-en} & \thead{nl-en} & \thead{pl-en} &  \thead{ru-en}  &  \thead{avg.}  \\
			\cmidrule(lr){1-1}
			\cmidrule(lr){2-10}
			\cmidrule(lr){11-11}
				GW     &  29.9	&52.9&	60.7&	32.7&	49.5&	57.6&	70.9&	57.7&	48.3& 47.0\\
			\textbf{\algname}  & \textbf{50.0}	& \textbf{57.7}&	\textbf{62.3}&	\textbf{54.4}&	\textbf{54.4}&	\textbf{61.0}&	\textbf{71.0}&	\textbf{60.5} & \textbf{54.1} & \textbf{53.0}\\
			\bottomrule
		\end{tabular}
	}
	\caption{P$@1$ for BLI on English and nine European languages: Bulgarian, Czech, Danish, Greek, Finnish, Hungarian, Dutch, Polish, and Russian.  The `avg.' shows P$@1$ averaged over all the eighteen BLI results. The proposed algorithm MBA outperforms GW when the bilingual mapping is learned between distant languages. }\label{table:ten_euro}
\end{table*}

\begin{table}[h]\centering
	\setlength{\tabcolsep}{5pt}
	{\small
		\centering
		\begin{tabular}{lrrrrrrr}
			\toprule
			\lhead{Method} &  \thead{en-ar} & \thead{en-hi}  &  \thead{en-tr} &  \thead{ar-en}  & \thead{hi-en} &  \thead{tr-en}  \\
			\cmidrule(lr){1-1}
			\cmidrule(lr){2-4}
			\cmidrule(lr){5-7}
			GW            &  27.4&	0.0&	40.9&	\textbf{41.0}&	0.0&	52.4\\
			\textbf{\algname}  &  \textbf{27.9} &	\textbf{25.1} &	\textbf{42.0} &	40.8 &	\textbf{28.9} &	\textbf{54.6} \\
			\bottomrule
		\end{tabular}
	}
	\caption{P$@1$ for BLI on English and three non-European languages (Arabic,  Hindi, and Turkish). MBA obtains significantly better results. 
	}\label{table:non_euro}
\end{table}

\section{Experiments}
We compare the proposed algorithm {\algname} with state-of-the-art GW alignment algorithm \citep{alvares18a} for the bilingual induction (BLI) task. Both the algorithms use second order statistics (word covariance matrices) to learn the word alignment between two languages. In our experimental setup, we first learn the word alignment between the source and the target languages and then compute cross-lingual mapping by solving the Procrustes problem (\ref{eq:procrustes}). 
For inference of nearest neighbors, we employ the cross-domain similarity local scaling (CSLS) similarity score \citep{conneau18a}. 
We report Precision$@1$ (P$@1$) as in \citep{alvares18a,artetxe18b} for the BLI task. 

We show results on the MUSE dataset \citep{conneau18a}, which consists of fastText monolingual embeddings for different languages \citep{bojanowski17} and dictionaries between several languages (but mostly with English). 
Following existing works \citep{artetxe18b,alvares18a,alaux19a}, the embeddings are normalized. 
The MUSE dataset provides predefined thirty test bilingual dictionaries between six European languages: English (en), German (de), Spanish (es), French (fr), Italian (it), and Portuguese (pt) on which we evaluate the methods. Additionally, we compute performance on the test dictionaries between English and twelve other languages: Arabic (ar), Bulgarian (bg), Czech (cs), Danish (da), Dutch (nl), Finnish (fi), Greek (el), Hindi (hi), Hungarian (hu), Polish (po), Russian (ru), and Turkish (tr). 
Following \citet{alvares18a}, we consider top $n = 20\,000$ most frequent words in the vocabulary set for all the languages during the training stage. The inference is performed on the the full vocabulary set. 

For GW, we use the original codes shared by \citet{alvares18a} and follow their recommendations on tuning the entropic regularization parameter and scaling of covariance matrices $\bC_\bX$ and $\bC_\bZ$. As a practical implementation of {\algname}, we incrementally increase $n$ starting from $1000$ to $20\,000$ every fixed-number of iterations.



We begin by discussing the results on  six close-by European languages in Table~\ref{table:six-euro}. We observe that both {\algname} and GW perform similarly when the languages are related. 
Hence, in the second set of experiments, we consider other European languages that are distant to English. 
We observe from Table \ref{table:ten_euro} that {\algname} outperforms GW, by an average BLI score of 6 points, in this challenging setting. 
Table~\ref{table:non_euro} reports results on language pairs involving English and three non-European languages. We again observe that the proposed algorithm MBA performs significantly better than GW. Overall, the experiments show the benefit of a geometric optimization framework.

\section{Conclusion}
Aligning the metric spaces of languages has a wide usage in cross-lingual applications. A popular approach in literature is the Gromov-Wasserstein (GW) alignment approach \citep{memoli11a,peyre16a,alvares18a}, which constructs a transport map by viewing the two embedding spaces as distributions. {In contrast, we have viewed} unsupervised bilingual word alignment as an instance of the more general unsupervised domain adaptation problem. {In particular, our formulation allows search over the space of doubly stochastic matrices and induces bi-directional mapping between the source and target words. Both are motivated solely from the language perspective. The Riemannian framework allows to exploit the geometry of the doubly stochastic manifold.} Empirically, we observe that the proposed algorithm {\algname} outperforms the GW algorithm for learning bilingual mapping \citep{alvares18a}, demonstrating the benefit of geometric optimization modeling. 




\bibliography{../multilingual-unsupervised-draft/main}
\bibliographystyle{acl_natbib}

\end{document}